\documentclass{article}
\usepackage{ijcai25}

\usepackage{times}
\usepackage{subfig}
\usepackage{soul}
\usepackage{url}
\usepackage[hidelinks,
            colorlinks=true,
            urlcolor=blue,
            citecolor=black,
            linkcolor=black,
            filecolor=black
           ]{hyperref}
\usepackage[utf8]{inputenc}
\usepackage{caption}
\usepackage{graphicx}
\usepackage{amsmath}
\usepackage{multirow}
\usepackage{amsthm}
\usepackage[authoryear,square,semicolon]{natbib}
\usepackage{booktabs}
\usepackage{algorithm}
\usepackage{algorithmic}
\usepackage[switch]{lineno}
\usepackage{longtable}

\title{Human Action CLIPs: Detecting AI-Generated Human Motion}

\vspace{0pt}

\author{
Matyas Bohacek$^{1,2}$
\and
Hany Farid$^3$
\affiliations
$^1$Google \ $^2$Stanford University \ $^3$University of California, Berkeley\\
\emails
\small{\texttt{maty@stanford.edu}},
\small{\texttt{hfarid@berkeley.edu}}
}

\begin{document}
\twocolumn[{%
\renewcommand\twocolumn[1][]{#1}%
\maketitle

\vspace{0pt}

\centering

\begin{tabular}{c@{\hspace{0.1cm}}c@{\hspace{0.1cm}}c@{\hspace{0.1cm}}c@{\hspace{0.1cm}}c@{\hspace{0.1cm}}c}
      \includegraphics[width=0.13\textwidth]{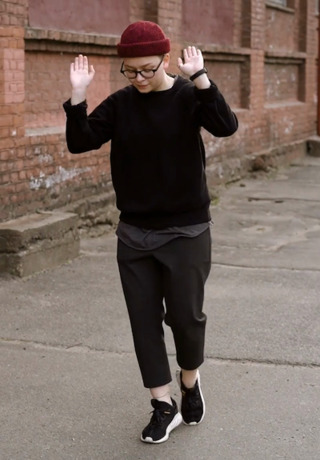} &
      \includegraphics[width=0.13\textwidth]{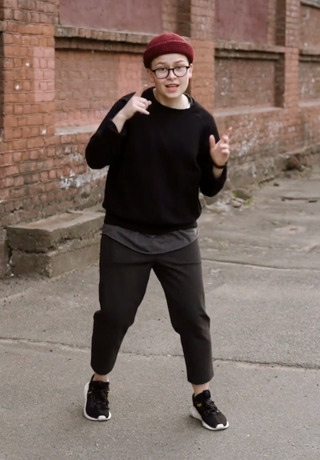} &
      \includegraphics[width=0.13\textwidth]{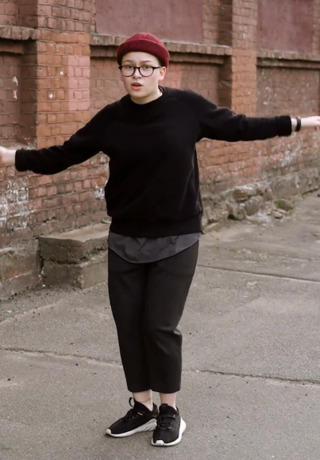} &
      \includegraphics[width=0.13\textwidth]{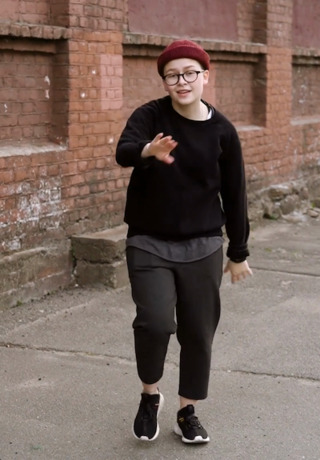} &
      \includegraphics[width=0.13\textwidth]{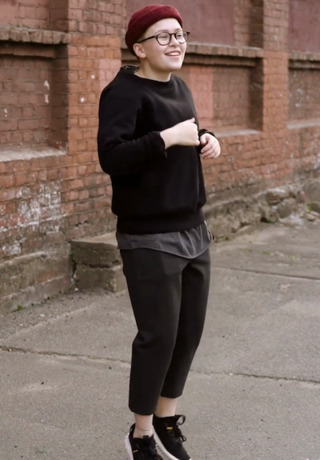} &
      \includegraphics[width=0.13\textwidth]{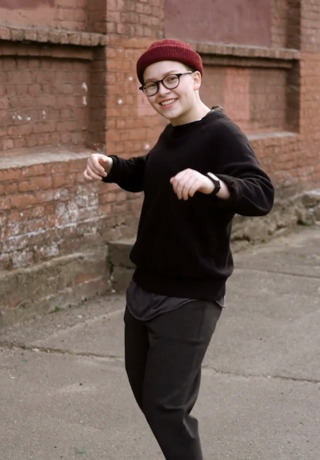} \\ 
      \includegraphics[width=0.13\textwidth]{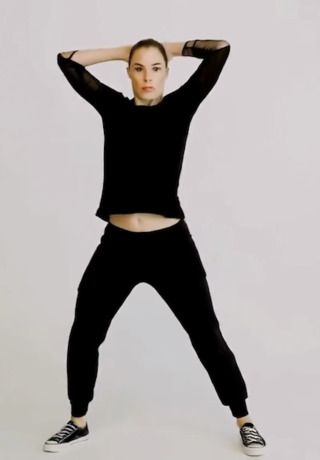} &
      \includegraphics[width=0.13\textwidth]{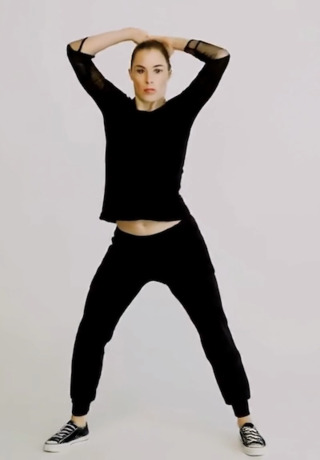} &
      \includegraphics[width=0.13\textwidth]{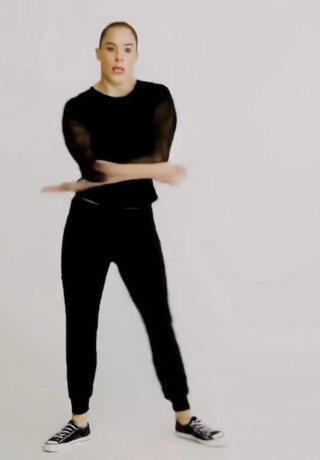} &
      \includegraphics[width=0.13\textwidth]{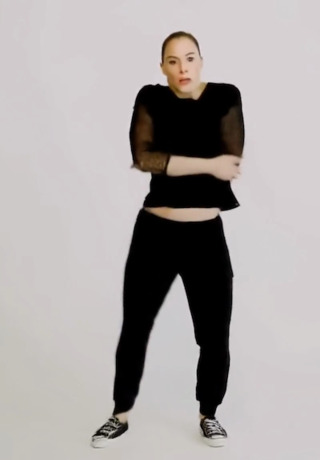} &
      \includegraphics[width=0.13\textwidth]{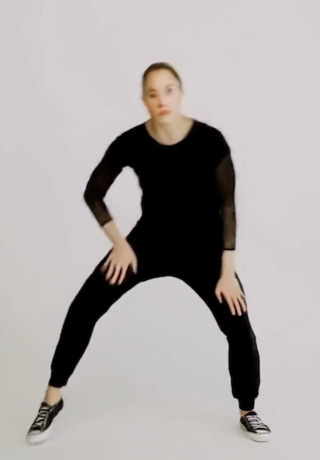} &
      \includegraphics[width=0.13\textwidth]{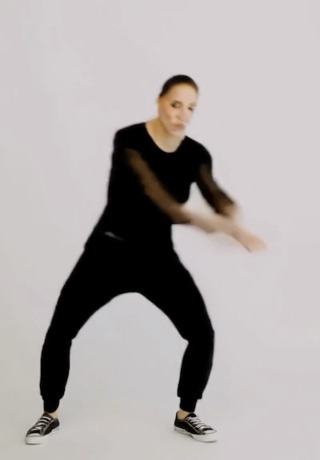} 
\end{tabular}

\begin{minipage}{0.81\textwidth}

\captionof{figure}{Real video frames sourced from Pexels (Top); AI-generated frames by Veo~\citep{veo} prompted with ``a person doing a dance move'' (Bottom). We describe how a low-dimensional CLIP embedding effectively and robustly distinguishes between real and AI-generated videos. \vspace{2em}}

\end{minipage}

\label{fig:teaser}

}]

\begin{abstract}
AI-generated video generation continues its journey through the uncanny valley to produce content that is increasingly perceptually indistinguishable from reality. To better protect individuals, organizations, and societies from its malicious applications, we describe an effective and robust technique for distinguishing real from AI-generated human motion using multi-modal semantic embeddings. Our method is robust to the types of laundering that typically confound more low- to mid-level approaches, including resolution and compression attacks. This method is evaluated against \textit{DeepAction}, a custom-built, open-sourced dataset of video clips with human actions generated by seven text-to-video AI models and matching real footage. The dataset is available under an academic license at \url{https://www.huggingface.co/datasets/faridlab/deepaction_v1}.


\end{abstract}

\section{Introduction}

Generating human motion in computer-graphics animation is notoriously difficult because of the complexity of human dynamics and kinematics~\citep{mcdonnell2012render,debarba2020plausibility,diel2021meta,ng2024audio} and because of the sensitivity of the human visual system to biological motion~\citep{neri1998seeing}. While motion capture has significantly improved the realism of complex human motion, gaps have remained. 

Generative AI, unlike earlier model-based animation, has emerged as an intriguing new genre for computer animation. While earlier versions of AI-generated video were things nightmares are made of (see, for example, ``Will Smith eating spaghetti''\footnote{\url{https://www.youtube.com/watch?v=XQr4Xklqzw8}}), recent advances have shown significant improvements in photo-realism and temporal consistency. 

A recent study found that AI-generated faces are nearly indistinguishable from real faces~\citep{nightingale2022ai}, and AI-generated voices are a close second in terms of naturalness and identity~\citep{barrington2024people}. There is reason to believe, therefore, that AI-generated videos may pass through the uncanny valley.

We have started to see AI-generated images weaponized in the form of child sexual abuse material, non-consensual sexual imagery, fraud, and as an accelerant to disinformation campaigns~\citep{farid2022creating}. There is no reason, therefore, to believe that AI-generated video will not follow suit. 

Video deepfakes fall into two broad categories: impersonation and text-to-video. Although there are several different incarnations of impersonation deepfakes, two of the most popular are lip-sync and face-swap deepfakes. In a face-swap deepfake, a person's face in an original video is replaced with another~\citep{nirkin2019fsgan}, and in a lip-sync deepfake, a person's mouth region is modified to be consistent with a new voice track~\citep{suwajanakorn2017}.

By contrast, text-to-video deepfakes are generated entirely from scratch to match a user-specified text prompt. They represent a natural evolution of text-to-image models (e.g.,~DALL-E, Firefly, Midjourney, etc.). Our focus is on detecting these text-to-video deepfakes, particularly those depicting human motion.

Motivated by, and building upon, earlier work, we describe a technique to detect AI-generated videos containing human motion. Our initial focus is on humans because these videos are of most concern when it comes to the harms enumerated above. Despite this focus, we will discuss why our approach is likely to generalize to other types of video content. We evaluate the efficacy of our approach on a diverse dataset of our creation consisting of real and matching (in terms of the human actions depicted) AI-generated videos from seven different generative-AI models. We demonstrate the robustness of our detection in the face of standard laundering attacks like resizing and transcoding, and evaluate its generalizability to previously unseen models. Our work contributes to this nascent literature with the:

\begin{figure*}[t]
    \centering
    \begin{minipage}[t]{0.48\textwidth}
        \centering
        {Real \vspace{0.1cm}} \\
        \includegraphics[width=0.32\textwidth]{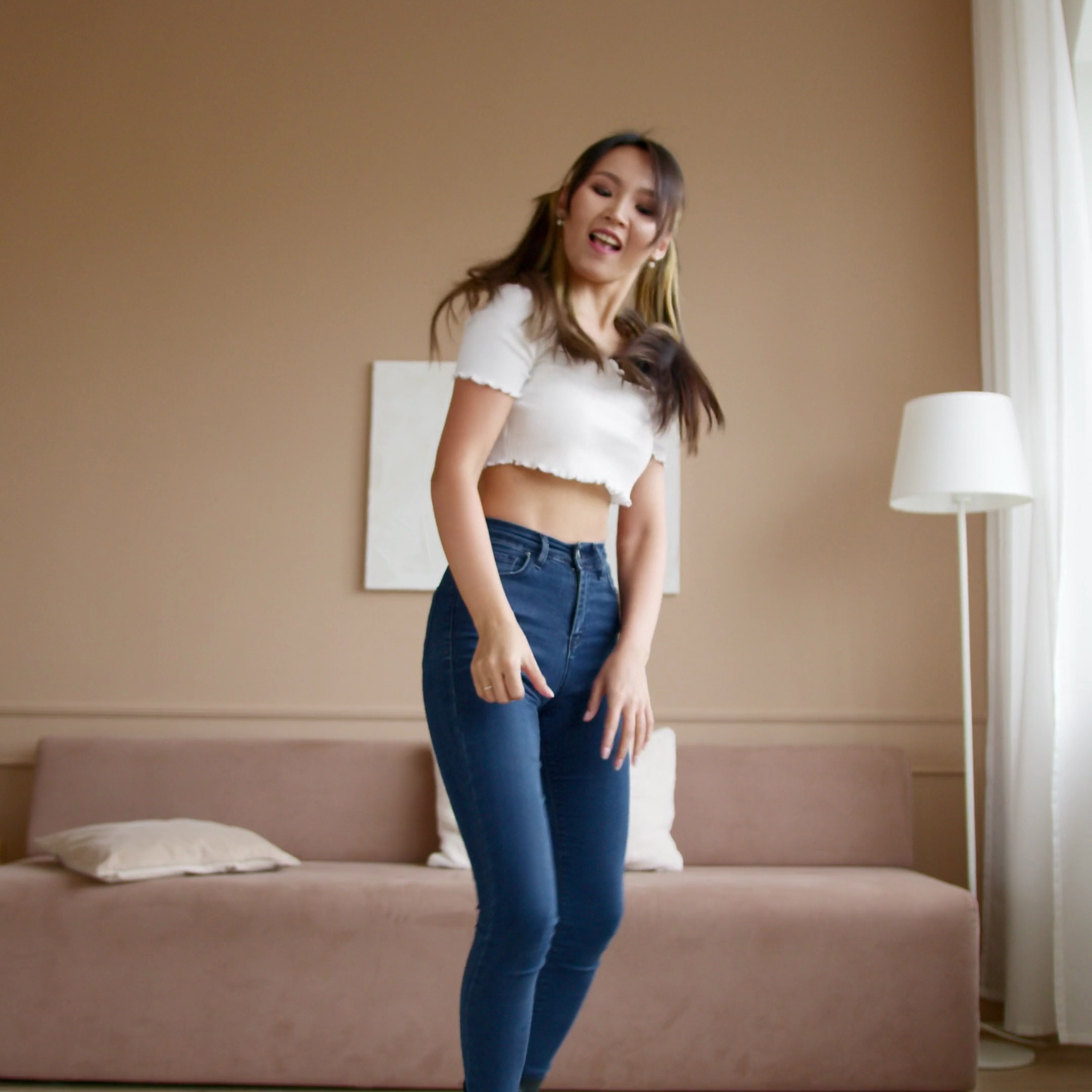}
        \includegraphics[width=0.32\textwidth]{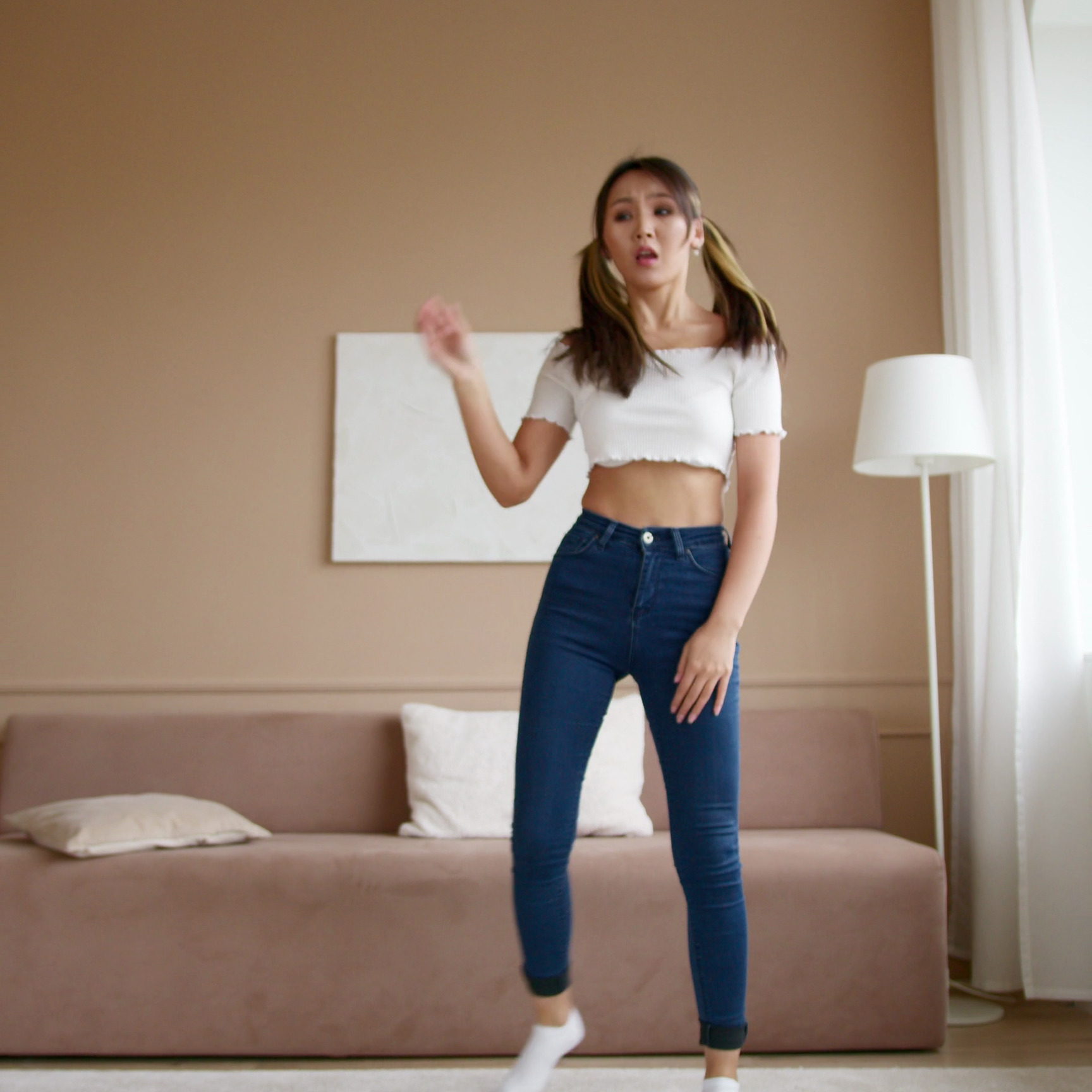}
        \includegraphics[width=0.32\textwidth]{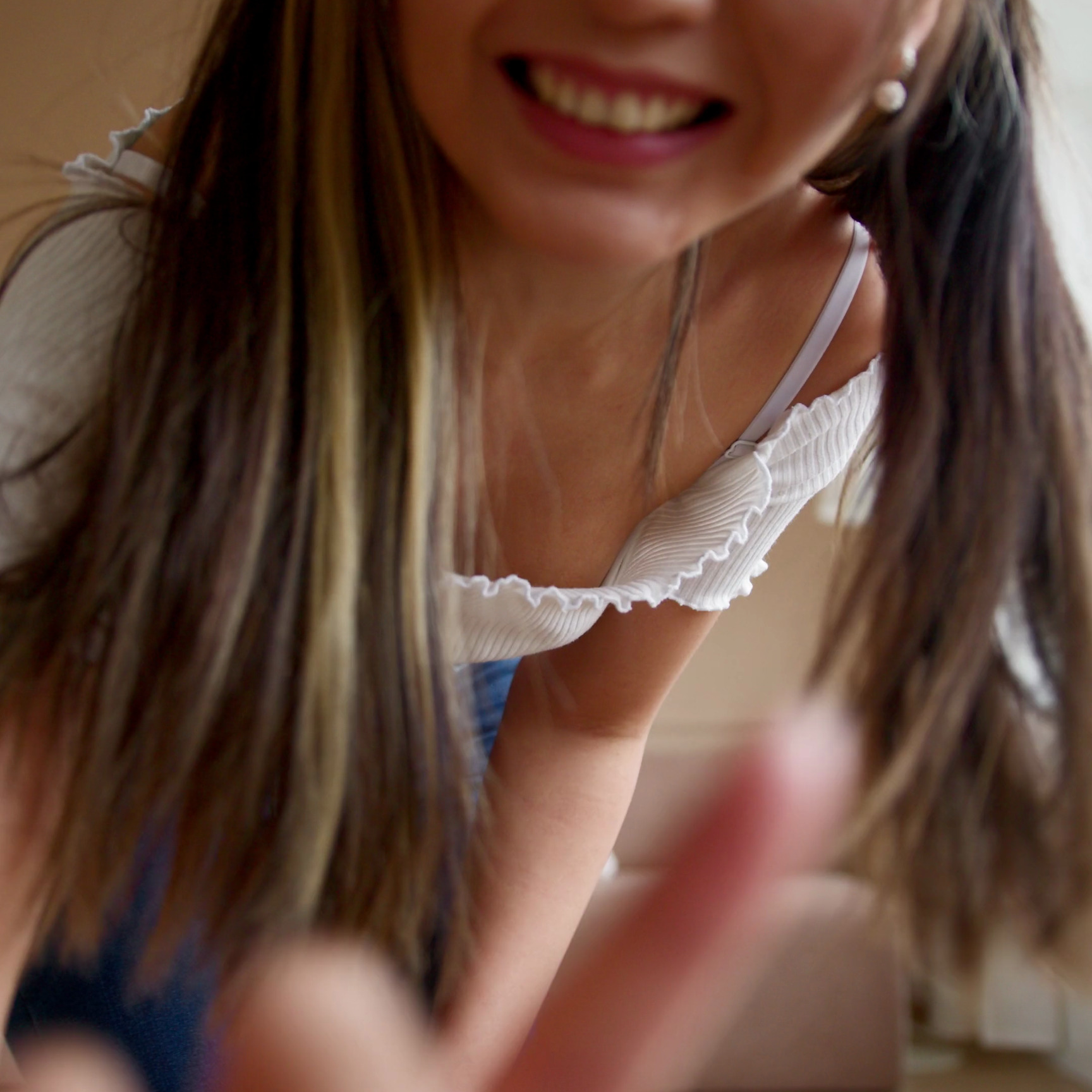}
    \end{minipage}%
    \hfill
    \begin{minipage}[t]{0.48\textwidth}
        \centering
        {BDAnimateDiffLightning \vspace{0.1cm}} \\
        \includegraphics[width=0.32\textwidth]{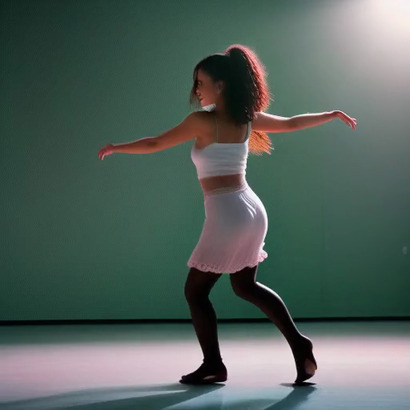}
        \includegraphics[width=0.32\textwidth]{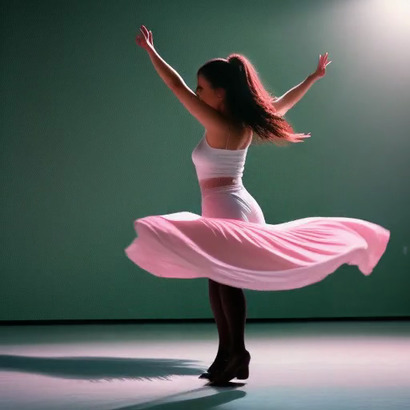}
        \includegraphics[width=0.32\textwidth]{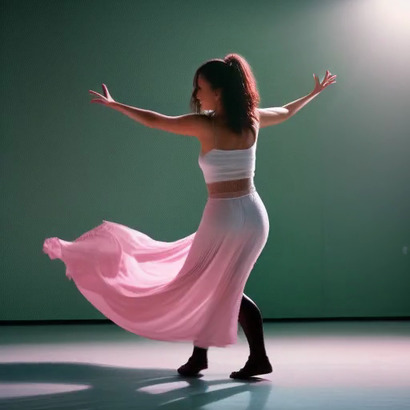}
    \end{minipage}%


    \begin{minipage}[t]{0.48\textwidth}
        \centering
        {\vspace{0.1cm} CogVideoX5B \vspace{0.1cm}} \\
        \includegraphics[width=0.32\textwidth]{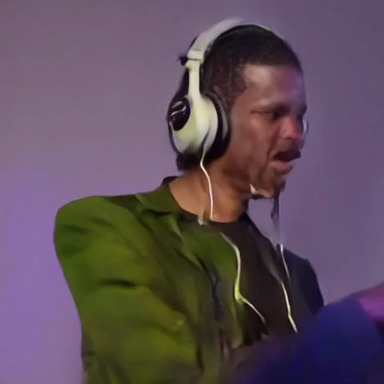}
        \includegraphics[width=0.32\textwidth]{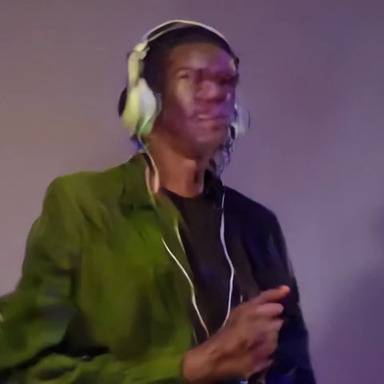}
        \includegraphics[width=0.32\textwidth]{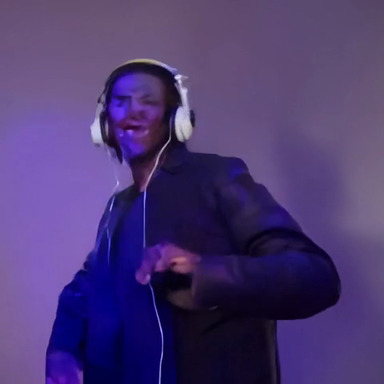}
    \end{minipage}
    \hfill
    \begin{minipage}[t]{0.48\textwidth}
        \centering
        {\vspace{0.1cm} Lumiere \vspace{0.1cm}} \\
        \includegraphics[width=0.32\textwidth]{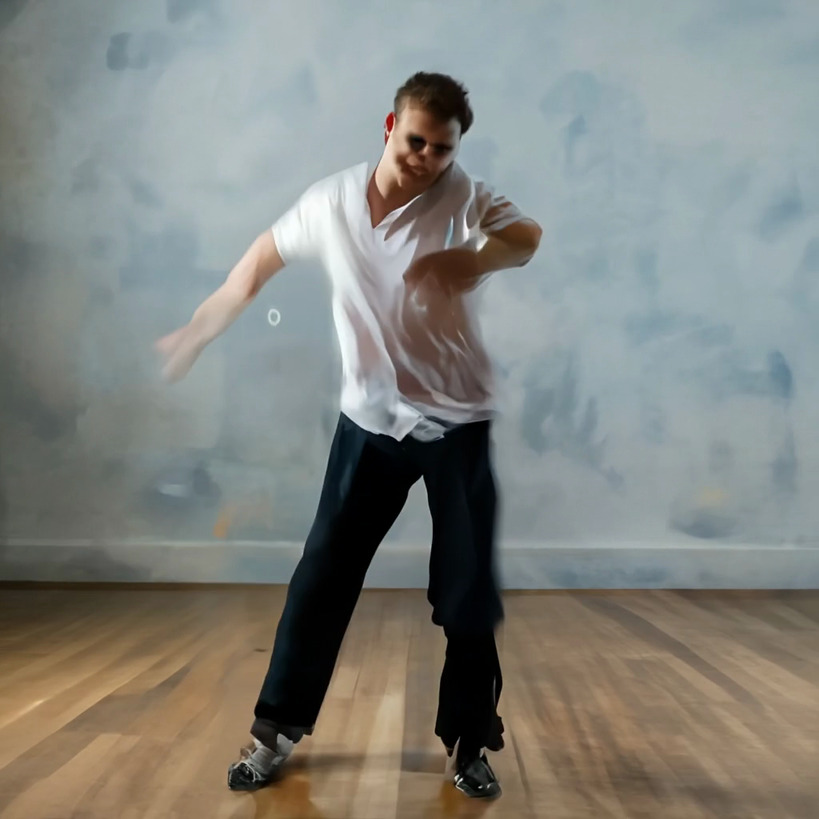}
        \includegraphics[width=0.32\textwidth]{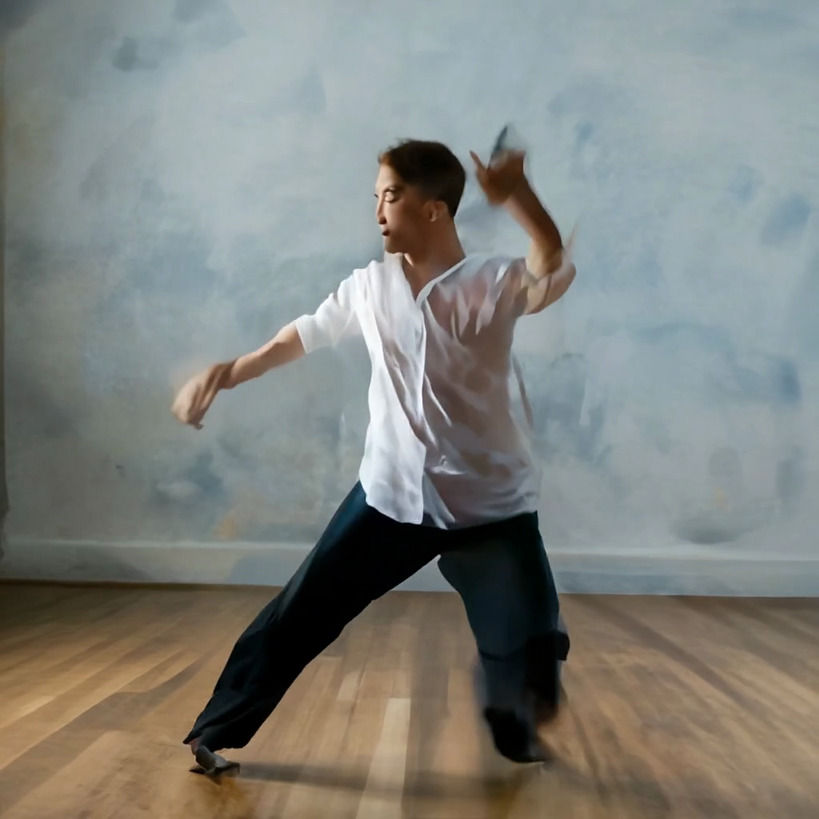}
        \includegraphics[width=0.32\textwidth]{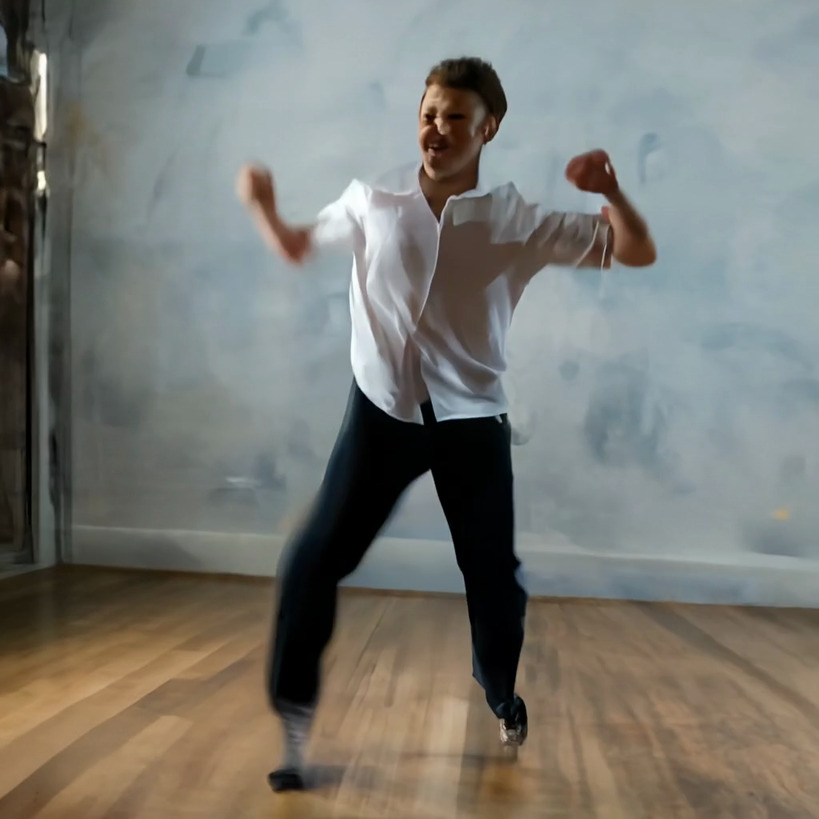}
    \end{minipage}%


    \begin{minipage}[t]{0.48\textwidth}
        \centering
        {\vspace{0.1cm} RunwayML \vspace{0.1cm}} \\
        \includegraphics[width=0.32\textwidth]{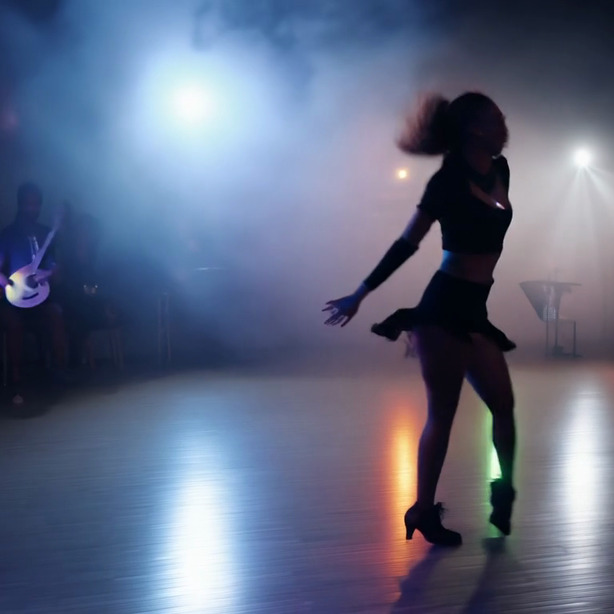}
        \includegraphics[width=0.32\textwidth]{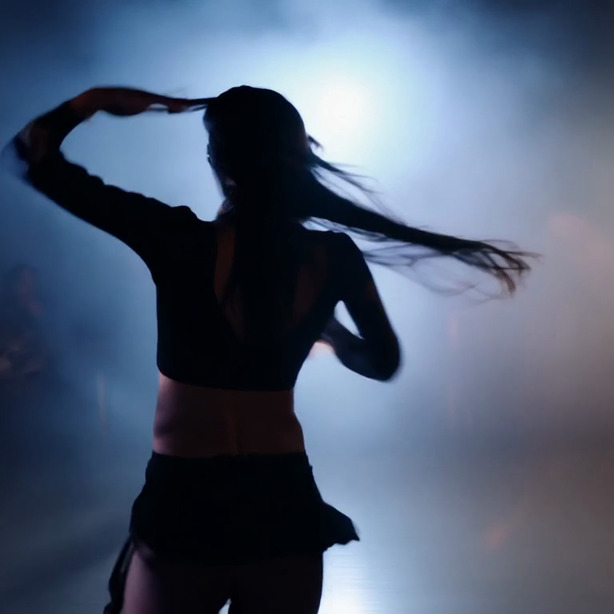}
        \includegraphics[width=0.32\textwidth]{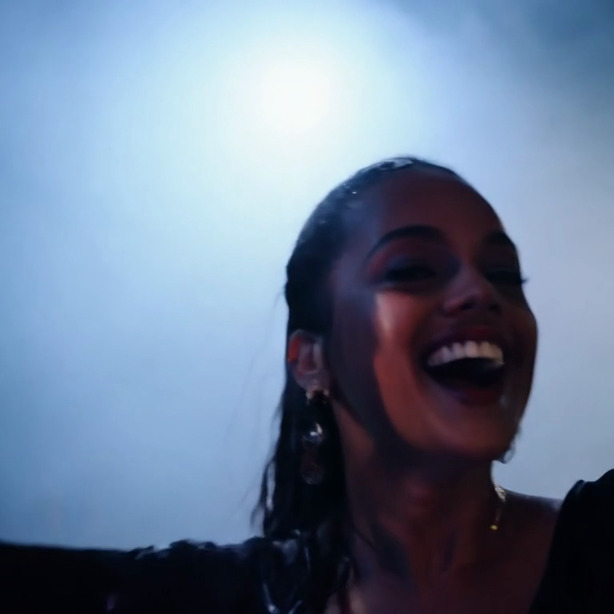}
    \end{minipage}%
    \hfill
    \begin{minipage}[t]{0.48\textwidth}
        \centering
        {\vspace{0.1cm} StableDiffusion \vspace{0.1cm}} \\
        \includegraphics[width=0.32\textwidth]{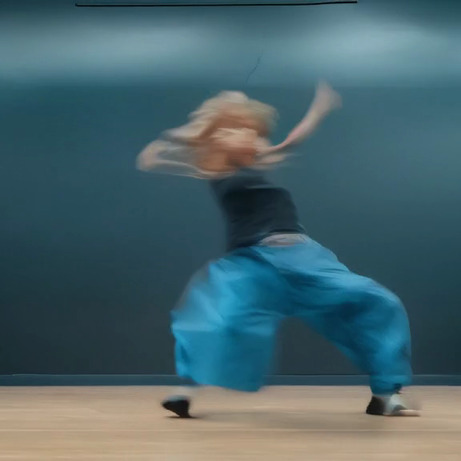}
        \includegraphics[width=0.32\textwidth]{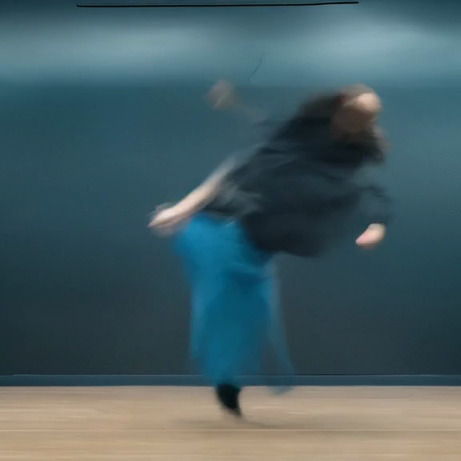}
        \includegraphics[width=0.32\textwidth]{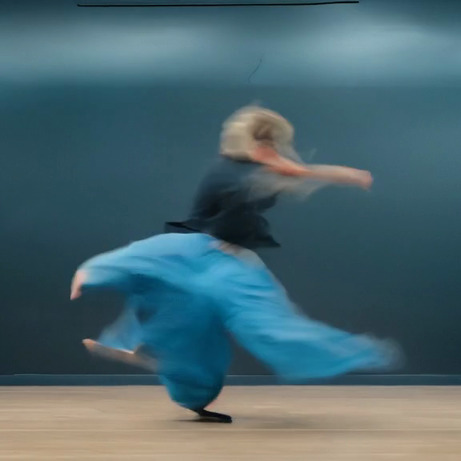}
    \end{minipage}    


    \begin{minipage}[t]{0.48\textwidth}
        \centering
        {\vspace{0.1cm} Veo \vspace{0.1cm}} \\
        \includegraphics[width=0.32\textwidth]{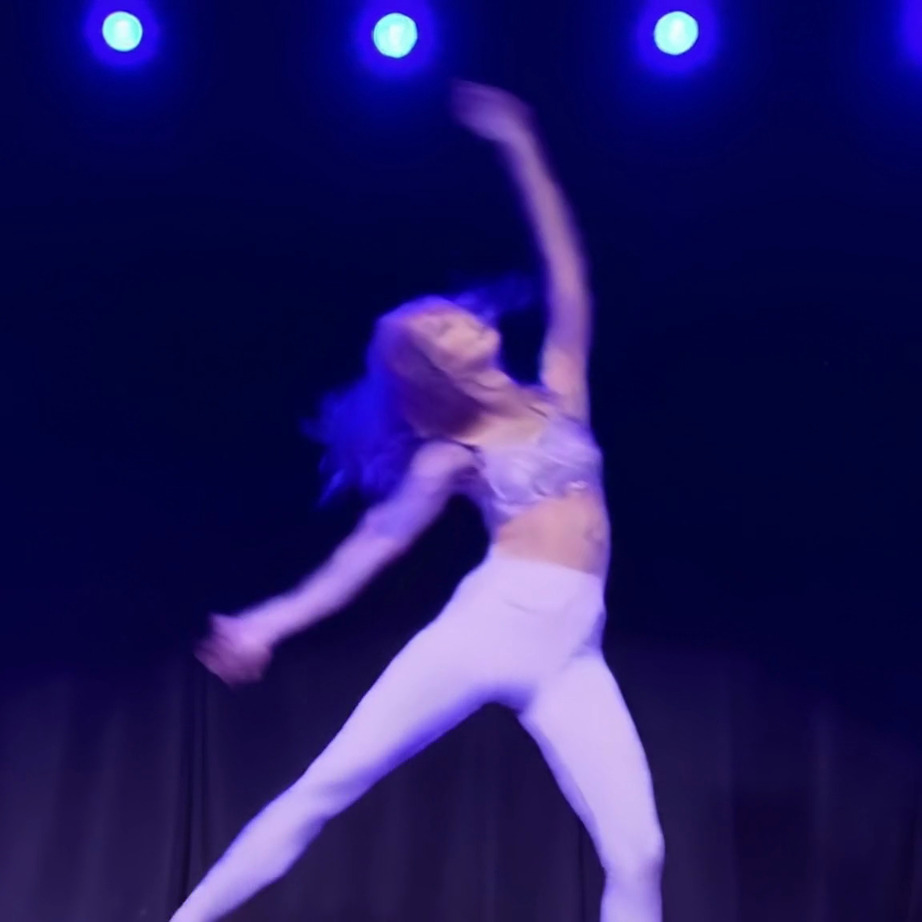}
        \includegraphics[width=0.32\textwidth]{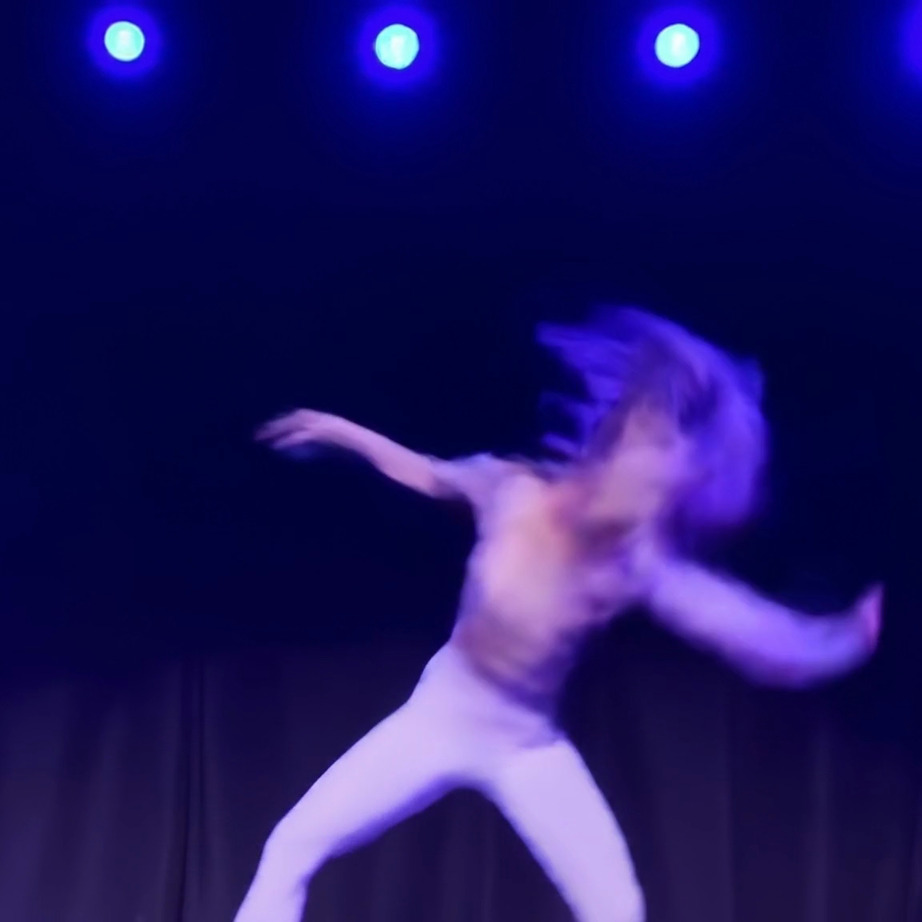}
        \includegraphics[width=0.32\textwidth]{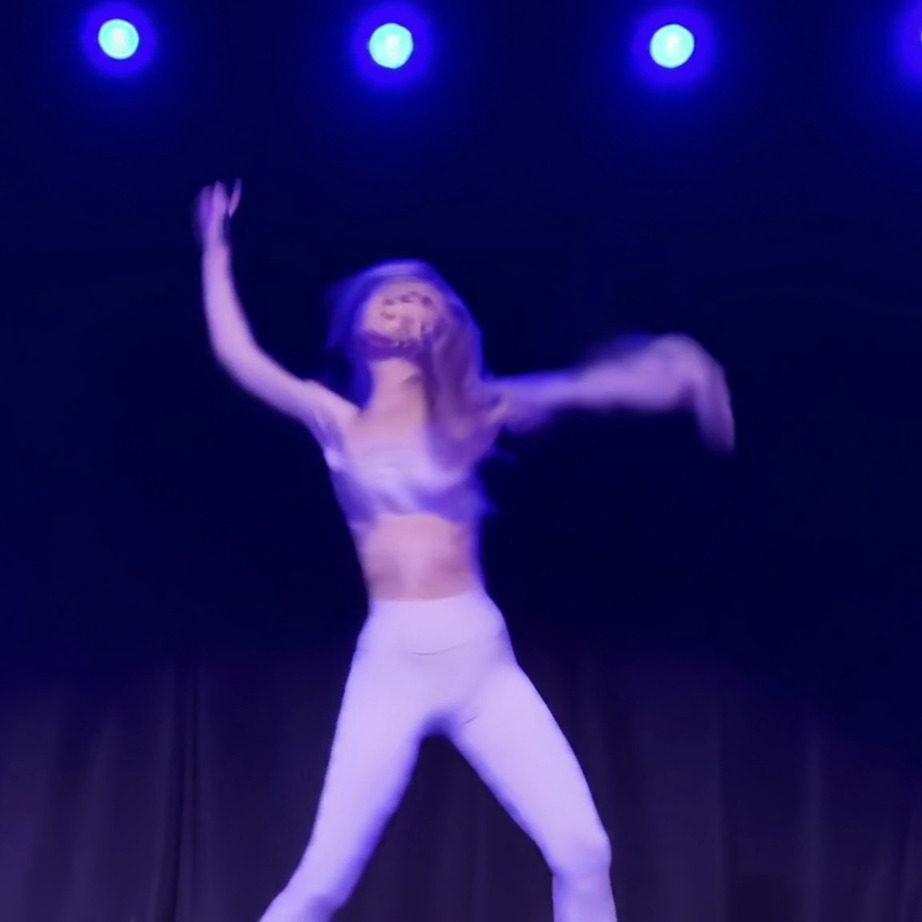}
    \end{minipage}%
    \hfill
    \begin{minipage}[t]{0.48\textwidth}
        \centering
        {\vspace{0.1cm} VideoPoet \vspace{0.1cm}} \\
        \includegraphics[width=0.32\textwidth]{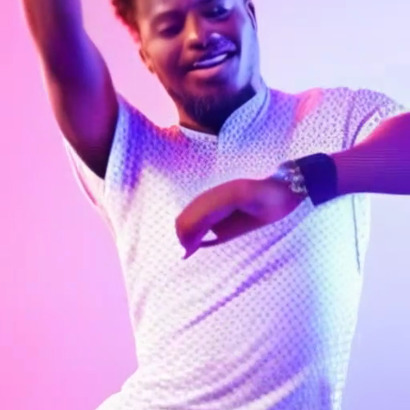}
        \includegraphics[width=0.32\textwidth]{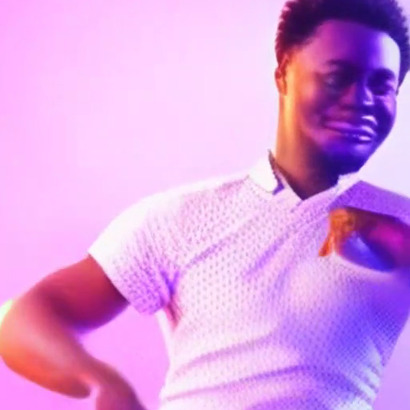}
        \includegraphics[width=0.32\textwidth]{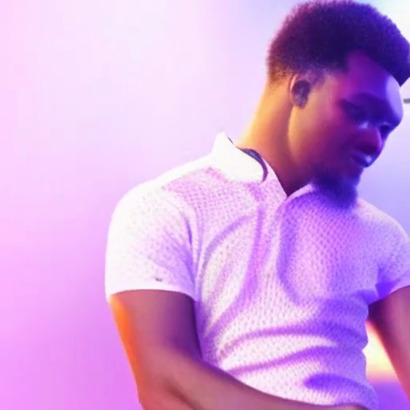}
    \end{minipage}

    \caption{Sample frames from one real and seven AI-generated videos prompted with ``a person dancing to music''.}
    \label{fig:deepaction}
\end{figure*}
\begin{enumerate}
    \item development of a task-specific CLIP embedding that outperforms generic CLIP embeddings (FT-CLIP);
    \item development of a new unsupervised forensic technique that requires no explicit training (frame-to-prompt);
    \item improvement in the generalizability of detection to previously unseen content and synthesis models;
    \item extension from image- to video-based analysis; and
    \item construction and release of \textit{DeepAction}, a new benchmark dataset of real and AI-generated human motion.
\end{enumerate}
%
%


\section{Related Work}

Identifying manipulated content (image, audio, video) can be partitioned into two broad categories: (1) active and (2) reactive. Active approaches involve inserting metadata or imperceptible watermarks at the time of synthesis to facilitate downstream detection~\citep{collomosse2024authenticity}. These approaches are appealing for their simplicity but are vulnerable to counter-attack in which the inserted credentials can be removed~\citep{voloshynovskiy2001attacks} (although the extraction and centralized storage of a distinct digital signature -- perceptual hash-- can be used to reattach credentials).

Reactive techniques -- operating in the absence of credentials -- fall into two basic approaches: (2a) learning-based, in which features that distinguish real from fake content are learned by a range of  machine-learning techniques, and (2b) artifact-based, in which a range of low-level (pixel-based) to high-level (semantic-based) features are explicitly extracted to distinguish between real and fake content~\citep{farid2022creating}.

There is a rich literature of techniques for detecting AI-generated images~\citep{farid2022creating} and a more nascent literature for detecting AI-generated voices~\citep{blue2022you,pianese2022deepfake,barrington2023single}. The literature for detecting AI-generated or AI-manipulated videos has primarily focused on face-swap deepfakes~\citep{agarwal2020detecting,nirkin2021deepfake,jia2022model}, and lip-sync deepfakes~\citep{bohavcek2022protecting,bohacek2024lost,datta2024exposing}.

Because text-to-video AI generation has only recently emerged as perceptually compelling, the literature on detecting these videos is more sparse. A recent example~\citep{vahdati2024beyond} leverages low-level features, but these tend to be vulnerable to compression artifacts, which in video -- unlike standard image JPEG compression -- are highly spatially and temporally variable. 

Another recent example~\citep{jia2024can} explores the potential of multi-modal large language models (LLMs) for detecting AI-generated faces.The authors prompt ChatGPT with an image and prompt like ``Tell me if this is an AI-generated image.'' This approach achieves an average accuracy of $75\%$ (as measured by area under the curve, AUC). Although not particularly accurate, what is intriguing about this approach is that it points to a potentially semantic-level reasoning. 

Recent studies took a more direct semantic approach by leveraging a contrastive language-image pretraining (CLIP) representation~\citep{radford2021learning}. In~\citep{cozzolino2024raising}, the authors extract a CLIP embedding from real and AI-generated images and with only a linear SVM achieve detection accuracy ranging from $85\%$ to $90\%$ depending on the amount of image post-processing. \nocite{de2024exploring}

In~\citep{khan2024clipping}, the authors also exploit CLIP embeddings along with a range of transfer learning strategies to achieve detection accuracy between $95\%$ and $98\%$; their classifiers show good but not perfect generalizability, achieving an accuracy between $86\%$ and $89\%$.

\nocite{smeu2024declip} 


\section{Dataset}
\label{subsec:datasets}

We generated $3{\small ,}100$ video clips from seven text-to-video AI models: BD AnimateDiff~\citep{lin2024animatediff}, CogVideoX-5B~\citep{yang2024cogvideox}, Lumiere\footnote{An earlier version of the Lumiere model was used which was not specifically trained or fine-tuned on human motion.}~\citep{bar2024lumiere}, RunwayML Gen3~\citep{runwayml}, Stable Diffusion Txt2Img+Img2Vid~\citep{blattmann2023stable}, Veo\footnote{A pre-release version of the Veo model was used.}~\citep{veo}, and VideoPoet~\citep{kondratyuk2023videopoet}. The default generation parameters were used for each model. These AI-generated videos are $297$ minutes in length constituting $254{\small ,}632$ video frames, Figure~\ref{fig:deepaction}.

These video clips depict $100$ distinct human actions and vary in length from $2$ to $10.7$ seconds, in resolution from $512 \times 512$ to $2048 \times 1152$ pixels, and in orientation (landscape and portrait). Each video was generated from a short prompt, which itself was generated by asking ChatGPT to create short descriptive prompts of human actions (see Appendix~\ref{app:prompts}).

We also curated a set of $100$ real videos from Pexels~\citep{pexels}, an open-source stock video database, matched to the human-action prompts described above. These real videos are $28$ minutes in length constituting $44{\small ,}475$ individual video frames. These videos were matched on general action so that the embedding representations (described next) between real and fake would not be based on semantic differences.

The resulting dataset, called \textit{DeepAction}, comprises eight video categories (seven AI-generated and one real) and is made publicly available.

\section{Methods}

We describe four CLIP-like, multi-modal embeddings and proceed with classification schemes used to distinguish the real from the fake.

\subsection{Embeddings}
\label{subsec:embeddings}

Multi-modal embedding models map an image and its corresponding descriptive caption into a shared vector space, allowing comparisons between these modalities. These embeddings have been found to be effective across many computer vision and natural language processing tasks~\citep{shen2021much,song2022clip}.

Each video in our dataset is represented as a sequence of video-frame embeddings, extracted using three off-the-shelf embedding models--CLIP~\citep{radford2021learning}, SigLIP~\citep{zhai2023sigmoid}, and JinaCLIP~\citep{koukounas2024jina}--as well as a custom fine-tuned CLIP model. 


%
%
\begin{enumerate}
    \item The {\bf CLIP} embedding model was trained on LAION-2B~\citep{schuhmann2022laion}. We used the smallest $512$-D baseline model ViT-B/32. 
    \item The {\bf SigLIP} embedding model was trained on WebLI~\citep{chen2022pali}, which is over $32\%$ larger than LAION-2B. Unlike CLIP's softmax-based contrastive loss, SigLIP uses a sigmoid loss. We used the $768$-D patch16-224 model variant.
    \item The {\bf JinaCLIP} embedding model was trained on LAION-400M~\citep{schuhmann2021laion} along with an additional set of $40$ text-pair datasets. While CLIP's training optimized for text-image representation alignment, JinaCLIP was trained to jointly optimize text-image and text-text representation alignment. We used the base $768$-D v1 model.
    \item We created a custom {\bf fine-tuned CLIP} (FT-CLIP) embedding model from the baseline CLIP model described above. We used the same fine-tuning methodology and hyperparameters as described in \citep{khan2024clipping}.  
\end{enumerate}

See Appendix~\label{app:exp_setup} for baseline model and training details.

\subsection{Classification}
\label{subsec:classification}

We deploy three different classification strategies that, leveraging the embeddings enumerated in the previous section, make a prediction of a video frame being real or fake. The first two supervised classifiers are based on a support vector machine (SVM), and the third unsupervised classifier is based on a simple cosine similarity between text and frame embedding. In each case, a video, represented as a sequence of frame embeddings, is classified as fake if a majority of the frames are classified as fake.

We intentionally take a simple approach here instead of leveraging heavier-weight classifiers so as to place emphasis on the power of the multi-modal embeddings.

\begin{enumerate}
    \item  A {\bf two-class SVM}~\citep{cortes1995support}, implemented in scikit-learn\footnote{\url{https://www.scikit-learn.org/stable/modules/generated/sklearn.svm.SVC.html}}, is used to classify video frames as real or fake, in which all seven text-to-video models are bundled into a single 'fake' class.
    %
    %
    \item A {\bf multi-class SVM}~\citep{cortes1995support}, also implemented in scikit-learn, classifies the source of each video frame across eight classes. One class represents real videos, and seven classes represent each of seven different text-to-video AI models.
    \item The previous classifiers follow a typical supervised learning approach in which the SVMs are trained on a subset of the video frames and evaluated on the remaining frames. In this third {\bf frame-to-prompt} approach, each video-frame embedding is compared -- through a simple cosine similarity -- to an embedding of one of two prompts (e.g.,~\textit{``a real image''} and \textit{``a fake image''}). A frame is classified by selecting the class (real/fake) with the largest cosine similarity. 
\end{enumerate}
%
%

\section{Results}
\label{sec:results}

We now describe a pair-wise embedding-classifier performance for discriminating between real and AI-generated videos, followed by an evaluation of the robustness in the face of standard laundering attacks, and generalizability to synthesis models not seen during classifier training.

For the CLIP, SigLIP, and JinaCLIP embeddings, our dataset is randomly split into an $80/20$ train/test partition. For the fine-tuned CLIP, the dataset is randomly split into a $40/40/20$ partition, where the first $40\%$ is used for CLIP fine-tuning, $40\%$ is used for model training, and $20\%$ is used for testing. The splits are determined at the action (prompt) level to ensure no overlap across partitions.

In each case, we report the mean frame- and video-level accuracy on the test set, averaged over five random train/test repetitions. Because our dataset is imbalanced, with significantly more fake than real videos, we under-sample the fake videos. Throughout, we report accuracy as a macro-average by evenly weighting the class accuracies.

\begin{table}[t]
\centering
\resizebox{0.47\textwidth}{!}{%
\begin{tabular}{|r|r|c|c||c|c|}
\hline
                         &                 & \multicolumn{2}{c||}{\textbf{Two-Class}} & \multicolumn{2}{c|}{\textbf{Multi-Class}} \\
\textbf{Embedding}       & \textbf{Kernel} & \textbf{Frame (\%)} & \textbf{Video (\%)} & \textbf{Frame (\%)} & \textbf{Video (\%)} \\ \hline \hline
\multirow{3}{*}{\textbf{CLIP}}     & \textbf{linear} & 84.3 & 97.0 & 87.4 & 98.2 \\ 
                                   & \textbf{RBF}    & 81.6 & 96.7 & 89.3 & 98.3 \\ 
                                   & \textbf{poly}   & 79.0 & 95.9 & 89.0 & 98.3 \\ \hline
\multirow{3}{*}{\textbf{SigLIP}}   & \textbf{linear} & 89.7 & 98.1 & \textbf{90.8} & \textbf{98.5} \\ 
                                   & \textbf{RBF}    & 84.4 & 97.1 & \textbf{91.5} & \textbf{98.5} \\ 
                                   & \textbf{poly}   & 83.0 & 96.9 & \textbf{90.9} & \textbf{98.4} \\ \hline
\multirow{3}{*}{\textbf{JinaCLIP}} & \textbf{linear} & 86.9 & 97.2 & 85.9 & 97.2 \\ 
                                   & \textbf{RBF}    & 78.1 & 96.1 & 87.3 & 97.9 \\ 
                                   & \textbf{poly}   & 76.1 & 95.3 & 87.0 & 97.8 \\ \hline
\multirow{3}{*}{\textbf{FT-CLIP}}  & \textbf{linear} & \textbf{90.7} & \textbf{98.5} & 82.8 & 97.6 \\
                                   & \textbf{RBF}    & \textbf{93.2} & \textbf{99.1} & 84.8 & 97.6 \\
                                   & \textbf{poly}   & \textbf{92.6} & \textbf{99.2} & 85.1 & 97.7 \\ \hline
\end{tabular}
}
\caption{Two-class and multi-class SVM classification accuracy (percent) for four different embeddings (CLIP, SigLIP, JinaCLIP, fine-tuned CLIP) and three different SVM kernels (linear, RBF, polynomial). Results are reported on a per video-frame and per video basis. Across all kernel functions, the fine-tuned CLIP provides the best performance for the two-class model, while SigLIP provides the best performance for the multi-class model.}
\label{tab:results-two-multi-class}
\end{table}
%
%


%
%
\begin{figure*}[t]  
    \centering
    \begin{minipage}[t]{0.49\textwidth}
        \vspace{0pt}
        \centering
        \begin{tabular}{c@{\hspace{0.1cm}}c}
            \includegraphics[width=0.42\textwidth]{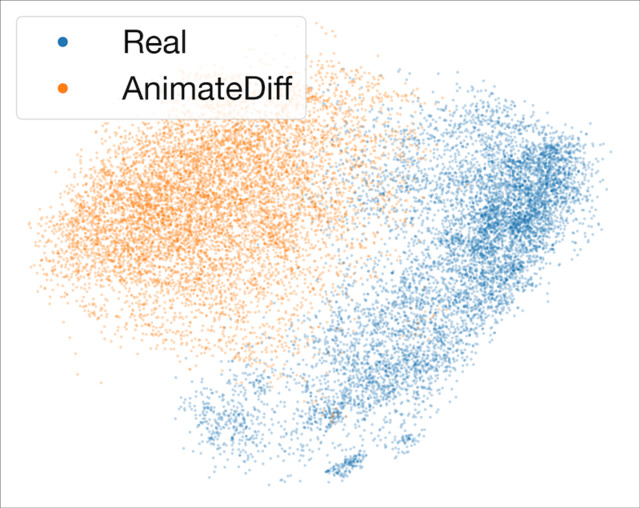} &
            \includegraphics[width=0.42\textwidth]{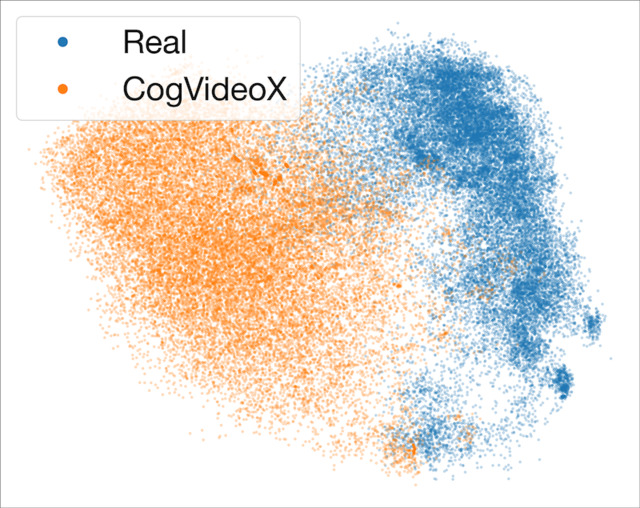} \\
            \includegraphics[width=0.42\textwidth]{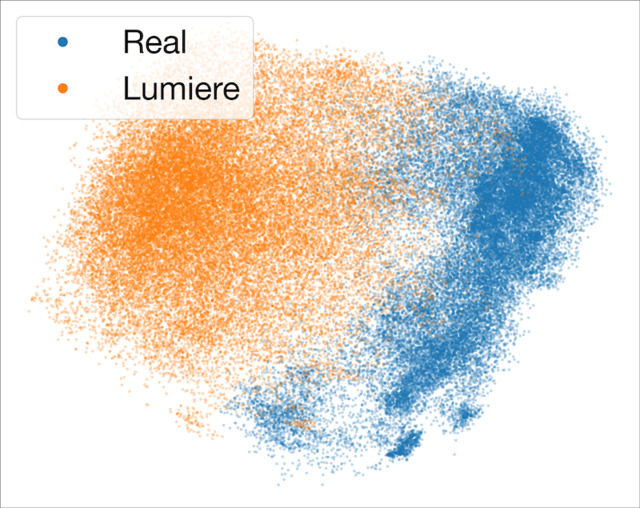} &
            \includegraphics[width=0.42\textwidth]{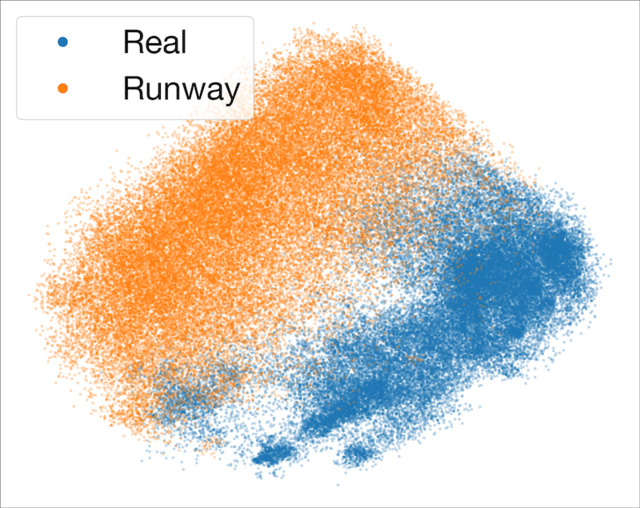} \\
            \includegraphics[width=0.42\textwidth]{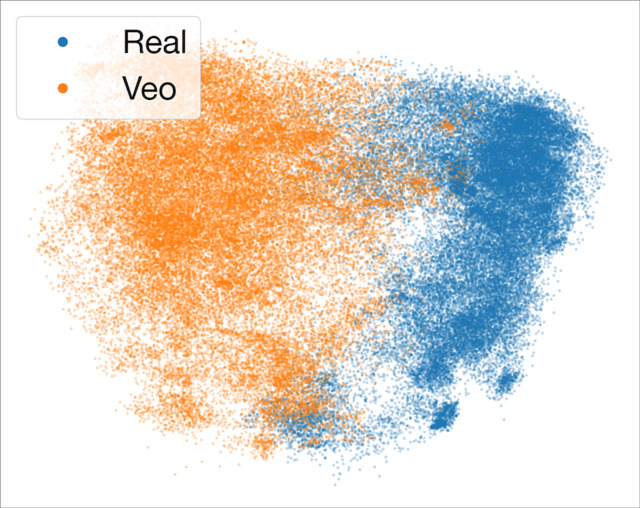} &
            \includegraphics[width=0.42\textwidth]{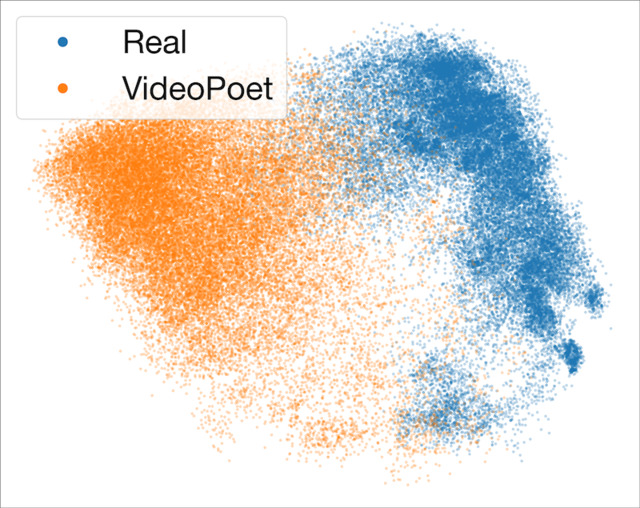}
        \end{tabular}
        \caption{Real (blue) vs fake (orange) fine-tuned CLIP embeddings. See Table~\ref{tab:results-two-multi-class}.}
        \label{fig:multi-class-viz-1}
    \end{minipage}
    \hfill
    \begin{minipage}[t]{0.49\textwidth}
        \vspace{0pt}
        \centering
        \begin{tabular}{c@{\hspace{0.1cm}}c}
            \includegraphics[width=0.42\textwidth]{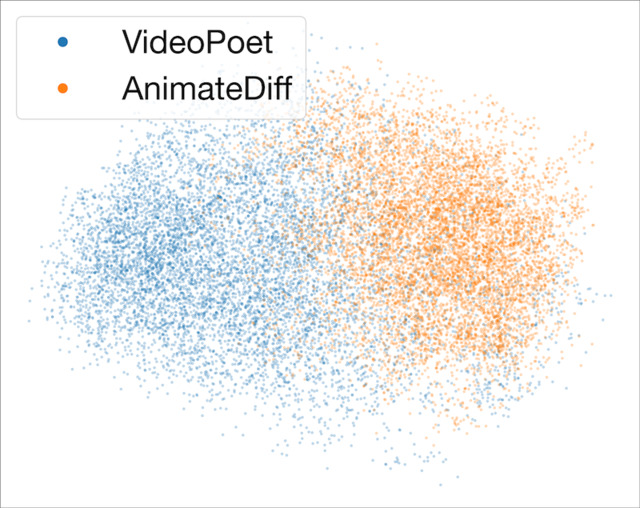} &
            \includegraphics[width=0.42\textwidth]{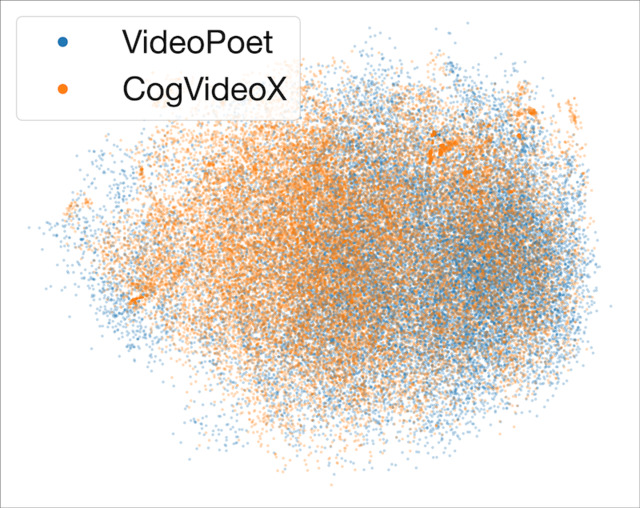} \\
            \includegraphics[width=0.42\textwidth]{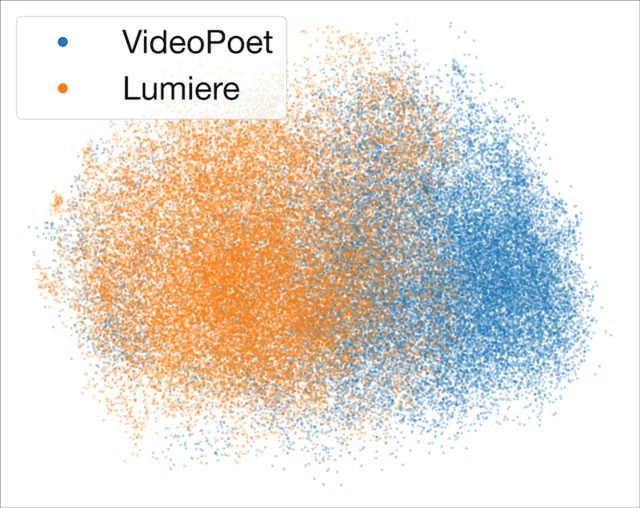} &
            \includegraphics[width=0.42\textwidth]{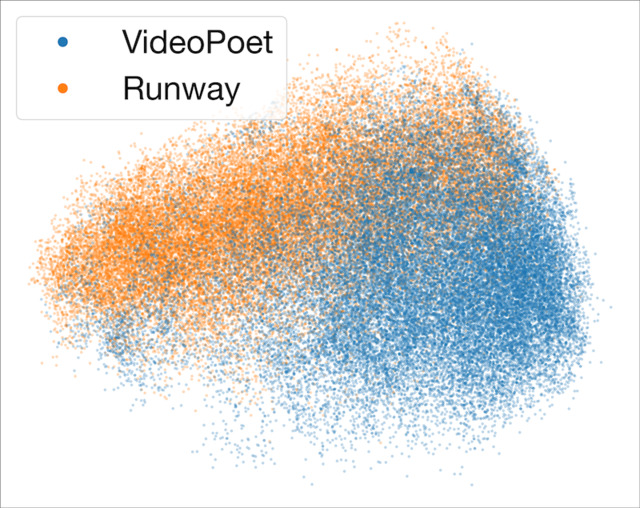} \\     
            \includegraphics[width=0.42\textwidth]{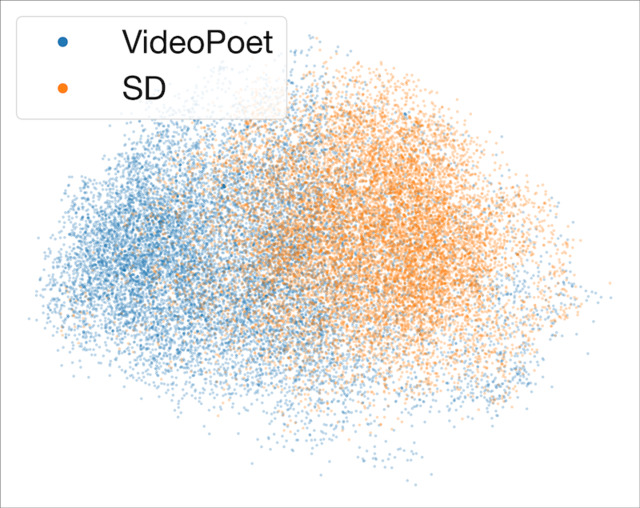} &
            \includegraphics[width=0.42\textwidth]{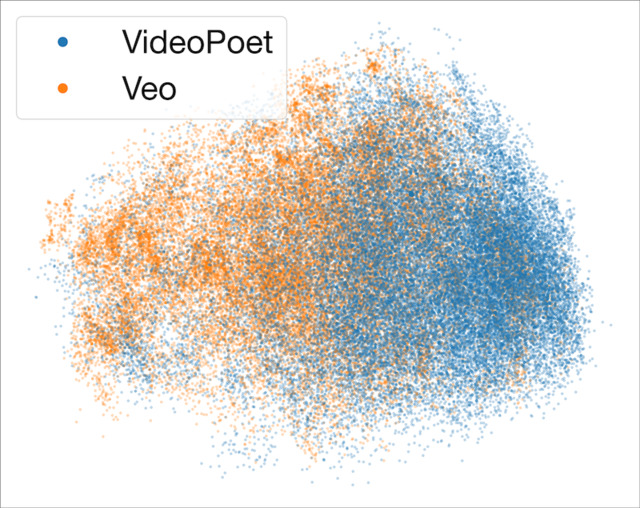} 
        \end{tabular}
        \caption{VideoPoet (blue) vs other models (orange) using fine-tuned CLIP embeddings.}
        \label{fig:multi-class-viz-2}
    \end{minipage}
\end{figure*}

{\bf Two-class.} Shown in the left portion of Table~\ref{tab:results-two-multi-class} is the frame- and video-level accuracy for the two-class SVMs with three different kernels: linear, radial basis function (RBF), and polynomial. 
    
At the frame level, macro-accuracy ranges from a low of $79.0\%$ to a high of $92.6\%$. For the CLIP embeddings (top three rows), the linear kernel is surprisingly more effective than the non-linear kernels. For the fine-tuned CLIP, the RBF kernel is the highest performer. Although performance is somewhat comparable across different embeddings, the fine-tuned CLIP offers the best performance. At the video level, accuracy across embeddings and classifiers is comparable, ranging from a low of $95.3\%$ to a high of $99.2\%$. 

For all embeddings and all classifiers, there is a slight fake bias in which fake content is correctly classified at a higher rate by, on average, $5.2$ percentage points. For example, for the best performing model (FT-CLIP with a poly kernel), the video-level accuracy for fake videos is $99.9\%$ as compared to $98.5\%$ for the real videos.

Shown in Figure~\ref{fig:multi-class-viz-1} is a visualization of seven pairwise, real-fake finetuned-CLIP embeddings, linearly reduced to a 2D subspace using PCA. Even in this reduced space, we see a good separation between the classes (the seven AI models are considered separately only for ease of visualization).

\vspace{0.25cm}

{\bf Multi-class.} Shown in the right portion of Table~\ref{tab:results-two-multi-class} is the frame- and video-level accuracy for the multi-class SVMs. At the frame level, accuracy slightly outperforms the two-class classifier, ranging from a low of $82.8\%$ to a high of $91.5\%$. Generally speaking, the non-linear kernels (RBF and polynomial) outperform the linear kernel, and unlike the two-class, the SigLIP embedding now outperforms the other embeddings across all kernel functions.

There is a slight bias across all seven AI models with the Veo model consistently misclassified. For example, for the best-performing model (SigLIP with an RBF kernel), the difference between the best (VideoPoet) and worst (Veo) performing inter-class accuracy is $100\%$ and $93.7\%$.

At the video level, accuracy across all embeddings and classifiers are comparable ranging from a low of $97.6\%$ to a high of $98.5\%$.

Shown in Figure~\ref{fig:multi-class-viz-2} is a PCA-based visualization of the pair-wise fine-tuned CLIP embeddings between one text-to-video model (VideoPoet) and each of six other text-to-video models. Even in this reduced space, we see a good separation between the different AI models (the six AI models are considered separately only for ease of visualization; other pair-wise models exhibit similar patterns).

\vspace{0.25cm}

{\bf Frame-to-prompt.} Shown in Table~\ref{tab:results-frame-to-prompt} is the frame- and video-level accuracy of frame-to-prompt classifiers (Section~\ref{subsec:classification}) for five different paired prompts: 
\begin{itemize}
    \item (P1) \textit{real photo vs. fake photo},
    \item (P2) \textit{real image vs. fake image},
    \item (P3) \textit{authentic image vs. AI-generated image},
    \item (P4) \textit{authentic photo vs. manipulated photo},
    \item (P5) \textit{authentic image vs. manipulated image}.
\end{itemize}

At the frame level, accuracy is generally significantly lower than the previous two- and multi-class SVMs, with a maximum accuracy of $95.2\%$ for the fine-tuned CLIP embedding and the P1 paired prompt \textit{authentic image vs. AI-generated image}. At the video level, accuracy for the same fine-tuned CLIP and P1 prompt improves slightly to $96.2\%$. This is perhaps not surprising since we expect these semantic embeddings to be highly correlated across a video.


As with the two-class SVM, we see a bias---only this time, towards real videos. For example, the video-level classification using FT-CLIP and prompt P1 correctly classifies $100\%$ of the real as compared to $92.4\%$ of the fake videos.

Compared to the best-performing two-class SVM with a video-level accuracy of $99.2\%$ (Table~\ref{tab:results-two-multi-class}), the best-performing frame-to-prompt underperforms by only $6.6$ percentage points. This is particularly impressive given that the frame-to-prompt approach requires no explicit training.

\begin{table}[t]
\centering
\begin{tabular}{|r|c|c|c|c|c|}
\hline
\multirow{2}{*}{\textbf{Embedding}} & \multicolumn{5}{c|}{\textbf{Frame (\%)}} \\ 
                         & \textbf{P1} & \textbf{P2} & \textbf{P3} & \textbf{P4} & \textbf{P5} \\ \hline \hline
\textbf{CLIP}            & 52.9 & 48.0 & 63.6 & 64.7 & 60.3 \\ \hline
\textbf{SigLIP}          & 49.0 & 47.8 & 49.1 & 49.1 & 47.2 \\ \hline
\textbf{JinaCLIP}        & 51.8 & 55.3 & 54.0 & 55.7 & 58.0 \\ \hline
\textbf{FT-CLIP}         & {\bf 95.2} & 91.0 & 83.1 & 78.3 & 79.9 \\ \hline
\hline
                         & \multicolumn{5}{c|}{\textbf{Video (\%)}} \\ 
                         & \textbf{P1} & \textbf{P2} & \textbf{P3} & \textbf{P4} & \textbf{P5} \\ \hline \hline
\textbf{CLIP}            & 53.5 & 44.2 & 62.0 & 63.7 & 58.4 \\ \hline
\textbf{SigLIP}          & 47.7 & 46.6 & 49.2 & 49.0 & 47.6 \\ \hline
\textbf{JinaCLIP}        & 56.7 & 55.6 & 56.2 & 59.9 & 60.5 \\ \hline
\textbf{FT-CLIP}         & {\bf 96.2} & 90.7 & 85.6 & 80.9 & 81.8 \\ \hline
\end{tabular}
\caption{Frame-to-prompt classification accuracy across different embeddings (CLIP, SigLIP, JinaCLIP,  fine-tuned CLIP) and five different prompt pairs: (P1) \textit{real photo vs. fake photo}; (P2) \textit{real image vs. fake image}; (P3) \textit{authentic image vs. AI-generated image}; (P4) \textit{authentic photo vs. manipulated photo}; and (P5) \textit{authentic image vs. manipulated image}. Results are reported on a per video-frame and per video basis.}
\label{tab:results-frame-to-prompt}
\end{table}

\subsection{Robustness}
\label{subsec:robustness}

Whether intentional or not, videos subjected to a forensic analysis often undergo laundering in the form of changes in resolution and compression. Techniques leveraging low-level features are often highly vulnerable to this type of laundering because even these simple modifications obliterate distinguishing characteristics. Because the multi-modal embeddings are designed to extract semantic-level meaning, we expect these features to be more resilient to laundering.

Shown in the left portion of Table~\ref{tab:results-robustness} are the frame- and video-level accuracies for the CLIP embedding and a two-class linear SVM for videos with progressively lower resolution (we quantify the change in resolution as a percentage because the original videos were of varying resolution, Section~\ref{subsec:datasets}). At the video level, accuracy remains relatively high for videos at $50\%$ or higher of their original resolution. This corresponds to an average resolution of $892 \times 355$.

Shown in the right portion of Table~\ref{tab:results-robustness} are the same frame- and video-level accuracies for videos with progressively lower compression (bits per second) compared to the original video compression. Here, accuracy remains high even for videos at $10\%$ of the original compression rate. This lowest compression corresponds to an average bit rate of $157$ kbps.

The loss in accuracy for lower-resolution videos suggests that there are some small-scale features that are important to distinguishing the real from the fake.

\begin{table}[t]
\centering
\resizebox{0.47\textwidth}{!}{%
\begin{tabular}{|r|c|c||c|c|}
\hline
                & \multicolumn{2}{c||}{\textbf{Resolution}} & \multicolumn{2}{c|}{\textbf{Compression}} \\
\textbf{amount} & \textbf{Frame (\%)} & \textbf{Video (\%)} & \textbf{Frame (\%)} & \textbf{Video (\%)} \\ \hline \hline
\textbf{100\%} & 83.1 & 96.9 & 83.1 & 96.9 \\ \hline
\textbf{75\%}  & 78.8 & 95.0 & 82.8 & 96.5 \\ \hline
\textbf{50\%}  & 72.5 & 93.8 & 81.8 & 96.5 \\ \hline
\textbf{25\%}  & 62.5 & 72.8 & 79.5 & 95.1 \\ \hline
\textbf{10\%}  & 49.5 & 51.0 & 73.7 & 92.3 \\ \hline
\end{tabular}
}
\caption{Accuracy for a two-class linear SVM and CLIP embeddings for videos at $100\%$, $75\%$, $50\%$, $25\%$, and $10\%$ of their original resolution (left) and compression ratio measured as bits per second (right). Results are reported on a per frame and per video basis.}
\label{tab:results-robustness}
\end{table}

\subsection{Generalizability}

Our frame-to-prompt classifier requires no training, which makes generalizability to new synthesis models more likely. Our two- and multi-class SVMs, however, do require training, and these types of supervised-learning models often struggle to generalize. To evaluate the generalizability of our SVMs, we performed a leave-one-out analysis in which two-class linear SVMs are trained on CLIP embeddings of six of the seven text-to-video models.

Shown in Table~\ref{tab:results-generalizability} are the frame- and video-level accuracies evaluated on just the real and left-out videos. When trained on all seven text-to-video models, accuracy is $96.9\%$ (Table~\ref{tab:results-two-multi-class}). By comparison, accuracy in this leave-one-out condition ranges from a low of $91.9\%$ to a high of $96.7\%$ with an average of $95.2\%$, showing that our approach generalizes well to previously unseen models.

\begin{table}[t]
\centering
\begin{tabular}{|r|c|c|}
\hline
\textbf{Leave-one-out}              & \textbf{Frame (\%)} & \textbf{Video (\%)} \\ \hline \hline
\textbf{BDAnimateDiffLightning}     & 82.3 & 94.9 \\ \hline
\textbf{CogVideoX5B}                & 82.6 & 95.7 \\ \hline
\textbf{Lumiere}                    & 79.2 & 91.9 \\ \hline
\textbf{RunwayML}                   & 81.4 & 94.8 \\ \hline
\textbf{StableDiffusion}            & 83.6 & 96.7 \\ \hline
\textbf{Veo}                        & 82.0 & 95.9 \\ \hline
\textbf{VideoPoet}                  & 82.6 & 96.2 \\ \hline
\end{tabular}
\caption{Average accuracy of a two-class linear SVM and CLIP embedding, where one text-to-video model is left out of the training set.}
\label{tab:results-generalizability}
\end{table}

\subsection{Non-Human Motion}

Having been trained on only videos containing human motion, we wondered if our two-class SVM would generalize to arbitrary text-to-video models. We evaluated our model on $100$ videos generated by Sora~\citep{sora} and RunwayML~\citep{runwayml} that did not contain any humans or human-motion. With the CLIP embeddings, the model correctly classified $97.1\%$ of these videos as AI-generated showing generalization to non-human motion and to one text-to-video model not seen in training (Sora). 

For the frame-to-prompt model, the highest video-level accuracy obtained was for the fine-tuned CLIP embedding and P3 prompt at an accuracy of $97.5\%$, on par with the human-motion (see Table~\ref{tab:results-frame-to-prompt}).

This result suggests that our model has not learned something specific to AI-generated human motion but instead has learned something distinct about AI-generated content. From a forensic perspective, this is highly desirable. 

We do not yet fully understand what specific properties of the multi-modal embeddings are distinct from AI-generated content. We speculate, however, that because generative-AI models rely on multi-modal embeddings to convert text prompts into images and videos, the extracted embeddings are distinct from those of real content. Another possibility is that generative-AI models may be prompted with distinct properties like level of descriptiveness, yielding more compact content as compared to real videos.

\subsection{Talking Heads}

We next wondered if our two-class SVM would generalize to a more constrained type of human motion in the form of face-swap and lip-sync deepfakes~\citep{farid2022creating}. We evaluated our model on $100$ real videos and $100$ deepfake videos from the DeepSpeak Dataset~\citep{barrington2024deepspeak}. These videos depict people sitting in front of their webcam responding to questions and prompts from which face-swap and lip-sync deepfakes were created. With the CLIP embeddings, the macro-average model accuracy is $55.8\%$ heavily biased to classifying content as fake at a rate of $93.1\%$ as compared to real at a rate of $15.2\%$. This is perhaps not surprising since, with the exception of the face, the content in these AI-manipulated videos is authentic.

For the frame-to-prompt model, the highest accuracy of $60.1\%$ is obtained from the fine-tuned CLIP embedding and P2 prompt, significantly lower than for the full-body, human-motion videos (see Table~\ref{tab:results-frame-to-prompt}).

Here, we do not see generalization with respect to the two-class model or frame-to-prompt. It remains to be seen if both approaches will improve with training in the case of the two-class model, and updating the fine-tuned CLIP with these types of deepfakes.

\subsection{CGI}

Lastly, we wondered if our two-class SVM would generalize to non-AI-generated video in the form of CGI. We evaluated the trained two-class SVM on videos from the GTA-Human dataset~\citep{cai2021playing}. We evaluated our model on $100$ GTA-Human videos (cropped around the human movement at a resolution of $640 \times 480$ pixels). With the CLIP embeddings, the model correctly classified only $39.1\%$ of these videos. Our model does not generalize to CGI.

Result for the frame-to-prompt model were mixed. Averaged across all prompts (P*), accuracy for the CLIP embedding was $76.1\%$, as compared to $15.0\%$ for JinaCLIP, and $48.9\%$ for FT-CLIP. 

Performance for SigLIP across all prompts was perfect at $100\%$. That is, the SigLIP embedding for these video frames is more similar to the ``AI-generated'', ``unrealistic'', or ``manipulated'' prompts than the ``authentic'' or ``realistic'' prompts. We don't fully understand why there is such a large discrepancy here across embedding, but it may be possible that SigLIP was exposed to CGI content during training.


\subsection{Comparison to Related Work}
\label{subsec:related-work}

Due to the lack of standardized benchmarks, comparing methods remains imperfect. Nevertheless, we provide a comparison to related work by comparing baseline accuracy and generalizability to unseen generative models. The latter is particularly important because true efficacy in the wild will be limited by the ability of detection models to generalize.

When trained on videos from all four text-to-video AI models, the method described in~\citep{vahdati2024beyond} attains an AUC in the range $98.5$ to $99.3$. When one model is left out, the AUC drops to between $67.1$ and $77.3$, revealing relatively poor generalizability.

When trained on videos from all models, our two-class linear SVM with CLIP embeddings attains an accuracy of $96.9\%$. When one model is left out, the accuracy drops to between $91.9\%$ and $96.7\%$. Although base-level accuracy is comparable, our method generalizes to videos from unseen models better than earlier approaches.

Although previous approaches have only focused on images, our frame- and video-level accuracy and generalizability are comparable or better than previous approaches~\citep{jia2024can,cozzolino2024raising,khan2024clipping} (Section~\ref{subsec:related-work}). Our frame-to-prompt is particularly attractive because it requires no explicit learning, making deployment relatively straightforward.


\section{Discussion}

When we first started to think about the problem of distinguishing real from AI-generated human motion, we thought to take a physics-based approach leveraging advances in 3D human-body modeling~\citep{loper2023smpl}. We hypothesized that by extracting 3D models of the human body, we could expose implausible dynamics and kinematics in AI-generated motion. We, however, quickly ran into obstacles and found it difficult to consistently and reliably extract 3D models in a wide range of human poses and levels of occlusion.

This led us to take a more learning-based approach. Wanting to avoid exploiting low-level features~\citep{vahdati2024beyond} vulnerable to laundering, we looked towards more semantic-level features, which, as we have shown, are more resilient to laundering. 

As with any authentication scheme, we must consider vulnerability to counter-attack by an adversary. Having already shown resilience to standard laundering, we will eventually need to consider a more sophisticated adversarial attack~\citep{carlini2020evading}. 

It has long been the goal -- and challenge -- of media forensics to extract semantic-level features that can distinguish real from fake or manipulated content. While we cannot say for sure that CLIP embeddings of the form explored here capture truly semantic properties, their resilience to resolution and quality and their generalizability suggest semantic-like representations. This is particularly the case for the surprisingly high performance achieved by our frame-to-prompt approach in which a video-frame embedding is simply compared to a pair of text embeddings of the form ``authentic image'' and ``AI-generated image.''

As text-to-video models advance in photorealism and computational efficiency~\citep{yu20244real}, complementary models for video-to-video generation and text-based video editing are also improving~\citep{bao2024vidu}. We should expect that the binary classification of real vs. fake will soon be complicated by hybrid videos that are partially AI-generated or AI-adjusted. Whether semantic methods, such as ours, can capture such subtleties remains to be seen. 


\section*{Ethical Statement}

Unlike the ethical considerations that should be weighed in the creation of generative-AI models (e.g.,~training data ownership and potential misuse in the form of the creation of non-consensual intimate imagery), we don't anticipate similar concerns with the development of techniques to detect AI-generated content. We recognize, however, that describing a detection technique could further enable the creation of more compelling deepfakes. We believe, however, that the benefit to the scientific field outweighs this risk, and to mitigate this risk we have chosen not to open-source our code.


\section*{Acknowledgements}

This work was done during the first author's (MB) internship at Google. The first author thanks his manager, Tony Chen, and co-host, Catherine Zhao, for their exceptional support, mentoring, and discussions related to this work; thanks to Prakash Hariramani, Sunita Verma, and Joshy Joseph for their support within leadership; and thanks for helpful discussions to: Amanda Walker, Athula Balachandran, Dara Bahri, David Hendon, Jess Gallegos, John Choi, Karen Guo, Nicholas Carlini, Nick Pezzoti, Stanley Wang, Sven Gowal, and Tom Hume. This work was supported by a gift from YouTube and funding from the University of California Noyce Initiative.


\bibliographystyle{named}
\bibliography{ijcai25}


\appendix


\section{Experimental Setup}

\subsection{CLIP}
The following model was used:

{\url{https://huggingface.co/openai/clip-vit-base-patch32}}

\subsection{SigLIP}
The following model was used:

{\url{https://huggingface.co/google/siglip-base-patch16-224}}

\subsection{JinaCLIP}
The following model was used: 

{\url{https://huggingface.co/jinaai/jina-clip-v1}}

\subsection{FT-CLIP}
The model was fine-tuned for one warm-up epoch at a learning rate of $10^{-5}$ and one regular epoch at a learning rate of $2 \cdot 10^{-3}$. An SGD optimizer with a cosine learning rate scheduler was employed for both epochs. The batch size was set to $16$ for training and $100$ for testing. The training set comprised two balanced sets of real and AI-generated frames, with captions constructed as \textit{"a \{REAL/FAKE\} image of \{ACTION PROMPT\}"}. Training was performed on an A100 GPU for approximately eight hours.

\setcounter{section}{0}
\onecolumn

\section{Video Generation Prompts}
\label{app:prompts}

\begin{longtable}{|p{0.45\textwidth}|p{0.45\textwidth}|}
\hline
A person walking through a park. & A person dancing in their living room. \\
\hline
A person doing yoga in their backyard. & A person cooking a meal in the kitchen. \\
\hline
A person playing fetch with their dog. & A person riding a bicycle down the street. \\
\hline
A person jogging on a sidewalk. & A person lifting weights at home. \\
\hline
A person practicing a musical instrument. & A person painting on a canvas. \\
\hline
A person doing push-ups in their living room. & A person stretching before a workout. \\
\hline
A person reading a book in a cozy chair. & A person writing in a journal. \\
\hline
A person meditating in a quiet room. & A person gardening in their backyard. \\
\hline
A person playing a board game with family. & A person folding laundry. \\
\hline
A person making a bed. & A person doing a puzzle. \\
\hline
A person brushing their hair. & A person tying their shoes. \\
\hline
A person washing dishes. & A person vacuuming the living room. \\
\hline
A person watering plants. & A person sewing on a sewing machine. \\
\hline
A person knitting on the couch. & A person ironing clothes. \\
\hline
A person mopping the floor. & A person baking cookies. \\
\hline
A person eating a meal at the table. & A person talking on the phone. \\
\hline
A person brushing their teeth. & A person playing with a pet. \\
\hline
A person working on a laptop. & A person watching TV. \\
\hline
A person drinking coffee on the porch. & A person taking a selfie. \\
\hline
A person organizing a bookshelf. & A person practicing calligraphy. \\
\hline
A person doing sit-ups. & A person practicing a dance move in front of a mirror. \\
\hline
A person applying makeup. & A person trimming a plant. \\
\hline
A person playing catch in the backyard. & A person skating in a driveway. \\
\hline
A person raking leaves. & A person cleaning windows. \\
\hline
A person decorating a cake. & A person unboxing a package. \\
\hline
A person eating popcorn while watching a movie. & A person putting together a DIY project. \\
\hline
A person reading a bedtime story. & A person practicing a speech. \\
\hline
A person blowing out birthday candles. & A person organizing their closet. \\
\hline
A person playing an online game. & A person having a video call. \\
\hline
A person building a model. & A person practicing origami. \\
\hline
A person doing jumping jacks. & A person dancing to music. \\
\hline
A person coloring in a coloring book. & A person taking a walk with a friend. \\
\hline
A person writing a letter. & A person enjoying a picnic. \\
\hline
A person birdwatching in the backyard. & A person making a smoothie. \\
\hline
A person cutting paper for a craft. & A person playing with building blocks. \\
\hline
A person wrapping a gift. & A person lighting a candle. \\
\hline
A person drawing a picture. & A person setting the table. \\
\hline
A person playing with a toy. & A person cleaning a mirror. \\
\hline
A person arranging flowers. & A person holding a pet. \\
\hline
A person organizing a drawer. & A person folding a paper airplane. \\
\hline
A person playing a card game. & A person practicing a yoga pose. \\
\hline
A person writing a grocery list. & A person doing a handstand against a wall. \\
\hline
A person making a sandcastle. & A person playing a sport in the backyard. \\
\hline
A person performing a simple magic trick. & A person showing a thumbs-up. \\
\hline
A person waving hello. & A person doing a dance move. \\
\hline
A person practicing a hobby. & A person clapping their hands. \\
\hline
A person jumping with joy. & A person making a funny face. \\
\hline
A person giving a high-five. & A person giving a thumbs-down. \\
\hline
A person walking up stairs. & A person walking down the stairs. \\
\hline
A person walking down a street. & A person jogging on a track. \\
\hline
\end{longtable}

\end{document}